\pdfoutput=1
\documentclass[10pt,twocolumn,letterpaper]{article}

\usepackage{wacv}
\usepackage{times}
\usepackage{epsfig}
\usepackage{graphicx}
\usepackage{amsmath}
\usepackage{amssymb}

\usepackage{caption}
\usepackage{subcaption}
\usepackage{cleveref}
\usepackage{multirow}
\usepackage{booktabs,tabularx}
\newcolumntype{Y}{>{\raggedright\arraybackslash}X} 
\usepackage{tabularx}
\usepackage{enumerate}
\usepackage{url}

\wacvfinalcopy 

\def\wacvPaperID{601} 

\ifwacvfinal\fi
\begin{document}
\newcommand{\MK}[1]{\textcolor{black}{#1}}
\newcommand{\PN}[1]{\textcolor{black}{#1}}

\title{PlotQA: Reasoning over Scientific Plots}

\author{Nitesh Methani \thanks{The first two authors have contributed equally} \hspace{1cm} Pritha Ganguly\textsuperscript{*} \hspace{1cm} Mitesh M. Khapra \hspace{1cm} Pratyush Kumar \\\\
Department of Computer Science and Engineering\\
Robert Bosch Centre for Data Science and AI (RBC-DSAI)\\
Indian Institute of Technology Madras, Chennai, India\\
{\tt\small \{nmethani, prithag, miteshk, pratyush\}@cse.iitm.ac.in}
}

\newcommand{\revision}[1][\textcolor{black}]{#1}
\newcommand{\delete}[1][\textcolor{b}]{#1}

\maketitle
\ifwacvfinal\thispagestyle{empty}\fi

\begin{abstract}
Existing synthetic datasets (FigureQA, DVQA) for reasoning over plots do not contain variability in data labels, real-valued data, or complex reasoning questions. 
Consequently, proposed models for these datasets do not fully address the challenge of reasoning over plots. \MK{In particular, they assume that the answer comes either from a small fixed size vocabulary or from a bounding box within the image. However, in practice, this is an unrealistic assumption because many questions require reasoning and thus have real-valued answers which appear neither in a small fixed size vocabulary nor in the image. In this work, we aim to bridge this gap between existing datasets and real-world plots}. 
Specifically, we propose PlotQA with \revision{28.9 million question-answer pairs} over \textcolor{black}{224,377 plots} on data from real-world sources and questions based on crowd-sourced question templates. 
Further, \revision{80.76\%} of the out-of-vocabulary (OOV) questions in PlotQA have answers that are not in a fixed vocabulary.
\MK{Analysis of existing models on PlotQA reveals that they cannot deal with OOV questions:  their overall accuracy on our dataset is in single digits. This is not surprising given that these models were not designed for such questions. As a step towards a more holistic model which can address fixed vocabulary as well as OOV questions, we propose a hybrid approach}: Specific questions are answered by choosing the answer from a fixed vocabulary or by extracting it from a predicted bounding box in the plot, while other questions are answered with a table question-answering engine which is fed with a structured table generated by detecting visual elements from the image. 
On the existing DVQA dataset, our model has an accuracy of 58\%, significantly improving on the highest reported accuracy of 46\%. 
On PlotQA, our model has an accuracy of \revision{22.52\%}, which is significantly better than state of the art models. 
\end{abstract}

\section{Introduction}

\begin{figure*}
\centering
\begin{subfigure}[t]{.32\textwidth}
  \centering
  \includegraphics[width=.9\linewidth]{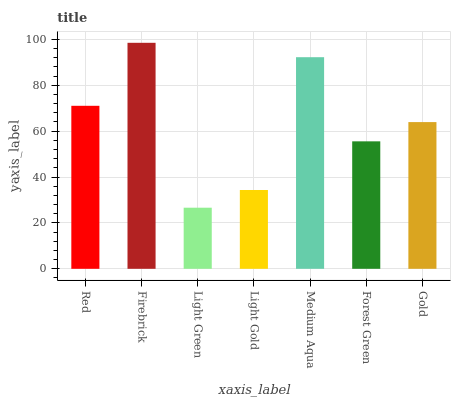}
\end{subfigure}%
\begin{subfigure}[t]{.32\textwidth}
  \centering
  \includegraphics[width=.9\linewidth]{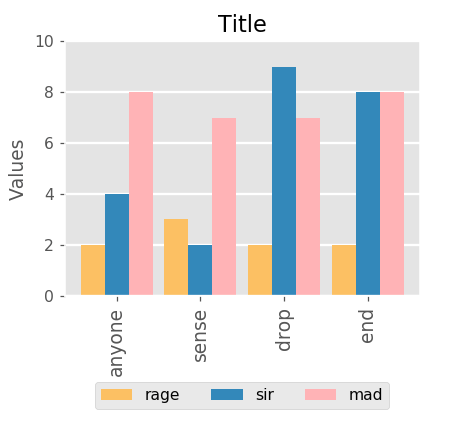}
\end{subfigure}
\begin{subfigure}[t]{.32\textwidth}
  \centering
  \includegraphics[width=1.02\linewidth]{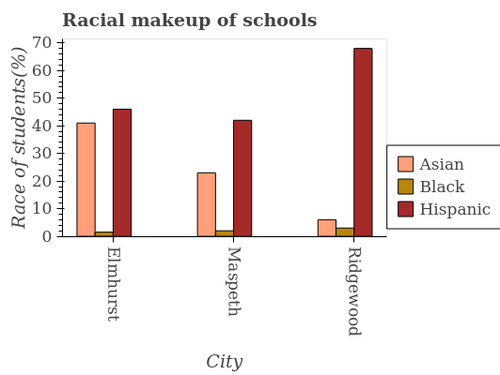}
\end{subfigure}
\begin{subfigure}[t]{.3\textwidth}
  \noindent
  \footnotesize
  Q: Is Light Green the minimum?\\
  A: $1$
  \caption{FigureQA}
  \label{fig:sample_plots_fqa}
\end{subfigure}
\begin{subfigure}[t]{.3\textwidth}
  \noindent 
  \footnotesize
  Q: What is the value of mad in drop?\\
  A: $7$
  \caption{DVQA}
  \label{fig:sample_plots_dvqa}
\end{subfigure}
\begin{subfigure}[t]{.3\textwidth}
  \noindent
  \footnotesize
  Q: What is the average number of Hispanic students in schools?
  A: $51.67$
  \caption{PlotQA}
  \label{fig:sample_plots_PlotQA}
\end{subfigure}
\caption{
A sample \{plot, question, answer\} triplet from FigureQA, DVQA, and PlotQA (our) datasets.
}
\label{fig:sample_plots}
\end{figure*}

Data plots such as bar charts, line graphs, scatter plots, etc. provide an efficient way of summarizing numerical information. 
Recently, in \cite{FQA, DVQA} two datasets containing plots and deep neural models for question answering over the generated plots have been proposed. In both the datasets, the plots are synthetically generated with data values and labels drawn from a custom set. In the FigureQA dataset \cite{FQA}, all questions are binary wherein answers are either Yes or No, (see Figure \ref{fig:sample_plots_fqa} for an example).  The DVQA dataset \cite{DVQA}, generalizes this to include questions which can be answered either by (a) fixed vocabulary of 1000 words, or (b) extracting text (such as tick labels) from the plot.  An example question could seek the numeric value represented by a bar of a specific label in a bar plot (see Figure \ref{fig:sample_plots_dvqa}).  Given that all data values in the DVQA dataset are chosen to be integers and from a fixed range, the answer to this question can be extracted from the appropriate tick label. 

\begin{table*}[ht!]
\begin{center}
\footnotesize
\begin{tabularx}{\textwidth}{|X|X|X|X|}
\hline
&\multicolumn{3}{|c|}{\textbf{Question Type}} \\ \cline{2-4}
\multicolumn{1}{|c|}{\textbf{Answer Type}} & \textbf{Structure} & \textbf{Data Retrieval} & \textbf{Reasoning} \\ \hline
\multirow{1}{*}{Yes/No}
 & \multicolumn{1}{l|}{\begin{tabular}[c]{@{}l@{}}Does the graph\\ contain grids?\end{tabular}} & \multicolumn{1}{l|}{\begin{tabular}[c]{@{}l@{}}Does the price of diesel in \\ Barbados monotonically\\ increase over the years?\end{tabular}} & \multicolumn{1}{l|}{\begin{tabular}[c]{@{}l@{}}Is the difference between the price of diesel in Angola in 2002 and 2004 \\greater than the difference between the price of diesel \\in Lebanon in 2002 and 2004?\end{tabular}} \\ \hline
\multirow{1}{*}{Fixed vocabulary}
 & \multicolumn{1}{l|}{\begin{tabular}[c]{@{}l@{}}How are the legend \\ labels stacked?\end{tabular}} & \multicolumn{1}{l|}{\begin{tabular}[c]{@{}l@{}}What is the label or title of \\ the X-axis ?\end{tabular}} & \multicolumn{1}{l|}{\begin{tabular}[c]{@{}l@{}}In how many years, is the price of diesel greater than 0.6 units?\end{tabular}} \\ \hline
\multirow{1}{*}{Open vocabulary}
 & - & \multicolumn{1}{l|}{\begin{tabular}[c]{@{}l@{}}What is the price of diesel in \\ Lebanon in the year 2008?\end{tabular}} & \multicolumn{1}{l|}{\begin{tabular}[c]{@{}l@{}}What is the ratio of the price of diesel in Lebanon in 2010 to that in 2014?\end{tabular}} \\ \hline
\end{tabularx}
\caption{\revision{Sample questions for 9 different question-answer types in PlotQA. The example questions are with respect to the plot in Figure \ref{fig:dataset_examples_line}. Note that there are no open vocabulary answers for \textit{Structural Understanding} questions.}}
\label{sample_qs_stats}
\end{center}
\end{table*}

\begin{table*}[]
\footnotesize
\begin{center}
\begin{tabular}{|l|c|c|c|c|c|c|c|c|}
\hline
\multicolumn{1}{|c|}{\textbf{Datasets}} & \textbf{\begin{tabular}[c]{@{}c@{}}\#Plot \\ types\end{tabular}} & \textbf{\begin{tabular}[c]{@{}c@{}}\#Plot\\ images\end{tabular}} & \textbf{\begin{tabular}[c]{@{}c@{}}\#QA \\ pairs\end{tabular}} & \textbf{\begin{tabular}[c]{@{}c@{}}Vocabulary \end{tabular}} & \textbf{\begin{tabular}[c]{@{}c@{}}Avg. question \\ length\end{tabular}} & \textbf{\#Templates} & \textbf{\begin{tabular}[c]{@{}c@{}}\#Unique \\ answers\end{tabular}} & \textbf{\begin{tabular}[c]{@{}c@{}}Open \\ vocab.\end{tabular}} \\ \hline
FigureQA & 4 & 180,000 & 2,388,698 & \begin{tabular}[c]{@{}c@{}}100 colours from\\ X11 colour set\end{tabular} & 7.5 & \begin{tabular}[c]{@{}c@{}}15 \\ (no variations)\end{tabular} & 2 & Not present \\ \hline
DVQA & 1 & 300,000 & 3,487,194 & \begin{tabular}[c]{@{}c@{}}1K nouns from\\ Brown corpus\end{tabular} & 12.30 & \begin{tabular}[c]{@{}c@{}}26 \\ (without paraphrasing)\end{tabular} & 1576 & Not present \\ \hline
PlotQA & 3 & 224,377 & 28,952,641 & \begin{tabular}[c]{@{}c@{}}Real-world  axes variables \\ and floating point numbers \end{tabular} & 43.54  & \begin{tabular}[c]{@{}c@{}}74 \\ (with paraphrasing)\end{tabular} & 5,701,618 & Present \\ \hline
\end{tabular}
\caption{\revision{Comparison between the existing datasets (FigureQA and DVQA) and our proposed dataset (PlotQA). }}
\label{tab:comparison_data_stats}
\end{center}
\end{table*}

While these datasets have initiated the research questions on plot reasoning, realistic questions over plots are much more complex. For instance, consider the question in Figure \ref{fig:sample_plots_PlotQA}, where we are to compute the average of floating point numbers represented by three bars of a color specified by the label.
The answer to this question is neither in a fixed vocabulary nor can it be extracted from the plot itself. 
Answering such questions requires a combination of perception, language understanding, and reasoning, and thus poses a significant challenge to existing systems.
Furthermore, this task is harder if the training set is not synthetic, but instead is sourced from real-world data with large variability in floating-point values, large diversity in axis and tick labels, and natural complexity in question templates.

To address this gap between existing datasets and real-world plots, we introduce the PlotQA\footnote{Dataset can be downloaded from \url{bit.ly/PlotQA}} dataset with \revision{28.9 million} question-answer pairs grounded over \textcolor{black}{224,377} plots.
PlotQA improves on existing datasets on three fronts.
First, roughly \revision{80.76\%} of the questions have answers which are not present in the plot or in a fixed vocabulary. 
Second, the plots are generated from data sourced from World Bank, government sites, etc., thereby having a large vocabulary of axis and tick labels, and a wide range in data values. 
Third, the questions are complex as they are generated based on 74 templates extracted from 7,000 crowd-sourced questions asked by workers on a sampled set of 1,400 plots. 
Questions are categorized into 9 (=3x3) cells based on the question type: `Structural Understanding', `Data Retrieval', or `Reasoning' and and the answer type: `Yes/No', `From Fixed Vocabulary', or `Out Of Vocabulary (OOV)' (see Table \ref{sample_qs_stats}). 

\MK{We first evaluate three state of the art models on PlotQA, \textit{viz.}, SAN-VQA\cite{SAN}, Bilinear attention network (BAN) \cite{DBLP:conf/nips/KimJZ18} and LoRRA \cite{singh2019TowardsVM}. Note that, by design none of these models are capable of answering OOV questions.
In particular, SAN-VQA and BAN treat plot reasoning as a classification task and expect the answer to lie in a small vocabulary whereas in our dataset the answer vocabulary is prohibitively large ($\sim$5M words). Similarly, LoRRA assumes that the answer is present in the image itself as a text and the task is to just extract this region containing the text followed by OCR (optical character recognition). Again, such a model will be unable to answer questions such as the one shown in Figure \ref{fig:sample_plots_PlotQA}, which form a significant segment of real-world use-cases and our dataset. As a result, these these models give an accuracy of less than 8\% on our dataset. On the other hand, existing models (in particular, SAN) perform well on questions with answers from a fixed vocabulary, 
which was the intended purpose of these models.}


\MK{Based on the above observations, we propose a hybrid model with a binary classifier which given a question decides if the answer would lie in a small top-$k$ vocabulary or if the answer is OOV. For the former, the question is passed through a classification pipeline which predicts a distribution over the top-$k$ vocabulary and selects the most probable answer. For the latter (arguably harder questions), we pass the question through a pipeline of four modules: Visual element detection, Optical character recognition, Extraction into a structured table, and Structured table question answering.}
This proposed hybrid model significantly outperforms the existing models and has an aggregate accuracy of \revision{22.52\%} on the PlotQA dataset.
We also evaluate our model on the DVQA dataset where it gives an accuracy of 58\%, improving on the best-reported result of SANDY \cite{DVQA} of 46\%.
In summary, we make two major contributions:\\
(1) We propose PlotQA dataset with plots on data sourced from the real-world and questions based on templates sourced from manually curated questions.
The dataset exposes the need to train models for questions that have answers from an Open Vocabulary. \\
(2) We propose a hybrid model with perception and QA modules for questions that have answers from an Open Vocabulary. This model gives the best performance not only on our dataset but also on the existing DVQA dataset.
\begin{figure*}
\centering
\begin{subfigure}[t]{.3\textwidth}
  \centering
  \includegraphics[width=.98\linewidth]{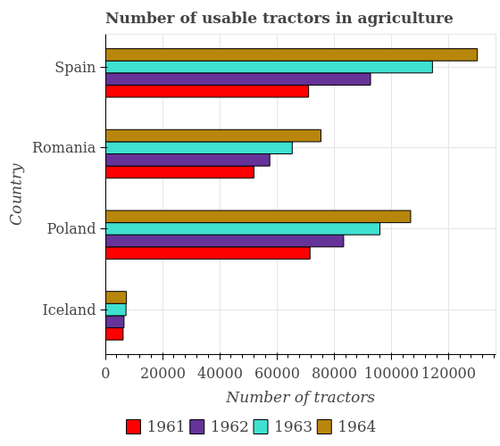}
  \caption{Horizontal bar graph}
  \label{fig:dataset_examples_hbar}
\end{subfigure}
\begin{subfigure}[t]{.3\textwidth}
  \centering
  \includegraphics[width=.95\linewidth]{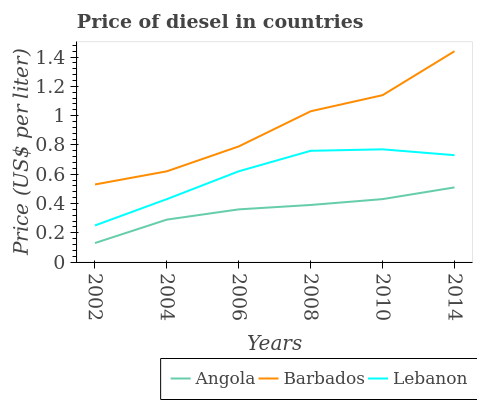}
  \caption{Line plot}
  \label{fig:dataset_examples_line}
\end{subfigure}
\begin{subfigure}[t]{.3\textwidth}
  \centering
  \includegraphics[width=1.05\linewidth]{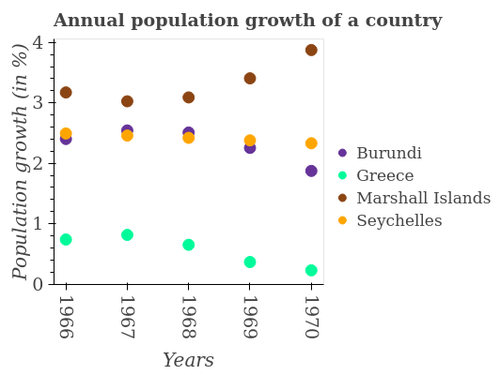}
  \caption{Dot-Line graph}
  \label{fig:dataset_examples_dotline}
\end{subfigure}
\caption{Sample plots of different types in the PlotQA dataset.}
\label{fig:dataset_examples}
\end{figure*}

\section{Related Work}

\noindent\textbf{Datasets:} 
Over the past few years several large scale datasets for Visual Question Answering have been released. 
These include datasets such as COCO-QA \cite{RenKZ15}, DAQUAR \cite{MalinowskiF14}, VQA \cite{AntolALMBZP15,GoyalKSBP17} which contain questions asked over natural images. 
On the other hand, datasets such as CLEVR \cite{JohnsonHMFZG17} and NVLR \cite{SuhrLYA17} contain complex reasoning based questions on synthetic images having 2D and 3D geometric objects.  
There are some datasets \cite{KembhaviSKSHF16,KembhaviSSCFH17} which contain questions asked over diagrams found in text books but these datasets are smaller and contain multiple-choice questions. 
FigureSeer \cite{SiegelHLDF16} is another dataset which contains images extracted from research papers but this is also a relatively small (60,000 images) dataset. 
Further, FigureSeer focuses on answering questions based on line plots as opposed to other types of plots such as bar charts, scatter plots, \textit{etc.} as seen in 
FigureQA \cite{FQA} and DVQA  \cite{DVQA}. \MK{There is also the recent TextVQA \cite{singh2019TowardsVM} dataset which contains questions which require models to read the text present in natural images. This dataset does not contain questions requiring numeric reasoning. Further, the answer is contained as a text in the image itself.}
Thus, \emph{no} existing dataset contains plot images with complex questions which require reasoning and have answers from an Open Vocabulary.

\noindent \textbf{Models:}
The availability of the above mentioned datasets has facilitated the development of complex end-to-end neural network based models (\cite{SAN}, \cite{LuYBP16}, \cite{YangHGDS16}, \cite{NohH16}, \cite{SantoroRBMPBL17}, \revision{\cite{DVQA}, \cite{singh2019TowardsVM}}).
These end-to-end networks contain (a) encoders to compute a representation for the image and the question, (b) attention mechanisms to focus on important parts of the question and image, (c) interaction components to capture the interactions between the question and the image, \revision{(d) OCR module to extract the image specific text and (e) a classification layer for selecting the answer either from a fixed vocabulary or from a OCR appended vocabulary. }
\section{The PlotQA dataset}

In this section, we describe the PlotQA dataset and the process to build it. 
Specifically, we discuss the four main stages, \textit{viz.}, (i) curating data such as year-wise rainfall statistics, country-wise mortality rates, \textit{etc.}, (ii) creating different types of plots with a variation in the number of elements, legend positions, fonts, \textit{etc.}, (iii) crowd-sourcing to generate questions, and (iv) extracting templates from the crowd-sourced questions and instantiating these templates using appropriate phrasing suggested by human annotators.

\subsection{Data Collection and Curation}

We considered online data sources such as World Bank Open Data
, Open Government Data
, Global Terrorism Database
, \textit{etc.} which contain statistics about various indicator variables such as fertility rate, rainfall, coal production, \textit{etc.} across years, countries, districts, \textit{etc.} 
We crawled data from these sources to extract different variables whose relations could then be plotted (for example, rainfall v/s years across countries, or movie v/s budget, or carbohydrates v/s food\_item). 
There are a total of 841 unique indicator variables (CO2 emission, Air Quality Index, Fertility Rate, Revenue generated, etc.) with 160 unique entities (cities, states, districts, countries, movies, food, etc.).
The data ranges from 1960 to 2016, though not all indicator variables have data items for all years.
The data contains positive integers, floating point values, percentages, and values on a linear scale. These values range from $0$ to $3.50\mathrm{e+}15$. 


\subsection{Plot Generation}
We included 3 different types of plots in this dataset, \textit{viz.}, bar plots, line plots, and scatter plots.
Within bar plots, we have grouped them by orientation as either horizontal or vertical. 
Figure \ref{fig:dataset_examples} shows one sample of each plot type. 
Each of these plot types can compactly represent 3-dimensional data. 
For instance, in Figure \ref{fig:dataset_examples_line}, the plot compares the \texttt{diesel prices} across \texttt{years} for different \texttt{countries}. 
To enable the development of supervised modules for various sub-tasks, we provide bounding box annotations for legend boxes, legend names, legend markers, axes titles, axes ticks, bars, lines, and title. 
By using different combinations of indicator variables and entities (years, countries, \textit{etc.}) we created a total of $224,377$ plots. 

To ensure variety in the plots, we randomly chose the following parameters: grid lines (present/absent), font size, notation used for tick labels (scientific-E notation or standard notation), line style (solid, dashed, dotted, dash-dot), marker styles for marking data points (asterisk, circle, diamond, square, triangle, inverted triangle), position of legends (bottom-left, bottom-centre, bottom-right, center-right, top-right), and colors for the lines and bars from a set of 73 colors. 
The number of discrete elements on the $x$-axis varies from 2 to 12 and the number of entries in the legend box varies from 1 to 4. 

\begin{table*}[ht!]
\footnotesize
\begin{center}
\begin{tabular}{|l|rrrr|rrr|rrr|}
\hline
\multirow{2}{*}{\textbf{\begin{tabular}[c]{@{}l@{}}Dataset\\Split\end{tabular}}} & \multicolumn{4}{c|}{\textbf{Plot Types}}                                                                                                              & \multicolumn{3}{c|}{\textbf{Question Types}}                                                                                      & \multicolumn{3}{c|}{\textbf{Answer Types}}                                                                                    \\ \cline{2-11} 
& \multicolumn{1}{l}{\textbf{vbar}} & \multicolumn{1}{l}{\textbf{hbar}} & \multicolumn{1}{l}{\textbf{line}} & \multicolumn{1}{l|}{\textbf{dot-line}} & \multicolumn{1}{l}{\textbf{Structural}} & \multicolumn{1}{l}{\textbf{Data-Retrieval}} & \multicolumn{1}{l|}{\textbf{Reasoning}} & \multicolumn{1}{l}{\textbf{Yes/No}} & \multicolumn{1}{l}{\textbf{Fixed vocab.}} & \multicolumn{1}{l|}{\textbf{Open vocab.}} \\ \hline
Train      & 52,463 & 52,700 & 25,897 & 26,010 & 871,782 & 2,784,041 & 16,593,656 & 784,115 & 3,095,774 & 16,369,590 \\ 
Validation & 11,249 & 11,292 & 5,547 & 5,571 & 186,994 & 599,573 & 3,574,081 & 167,871 & 600,424 & 3,592,353 \\ 
Test       & 11,242 & 11,292 & 5,549 & 5,574 & 186,763 & 596,359 & 3,559,392 & 167,727 & 667,742 & 3,507,045 \\ \hline
\end{tabular}
\caption{\PN{Detailed Statistics for different splits of the PlotQA dataset.}}
\label{tab:detailed_stats}
\end{center}
\end{table*}


\begin{table}[]
\footnotesize
\begin{center}
\begin{tabular}{|l|r|r|r|}
\hline
&\multicolumn{3}{|c|}{\textbf{Question (Q) Type}} \\ \cline{2-4}
\multicolumn{1}{|c|}{\textbf{Answer (A) Type}} & \multicolumn{1}{c|}{\textbf{Structure}} & \multicolumn{1}{c|}{\textbf{Data Retrieval}} & \multicolumn{1}{c|}{\textbf{Reasoning}} \\ \hline
Yes/No & 36.99\% & 5.19\% & 2.05\% \\
Fixed  vocabulary & 63.01\% & 18.52\% & 15.92\% \\
Open vocabulary & 0.00\% & 76.29\% & 82.03\% \\ \hline
\end{tabular}
\caption{\PN{Overall distribution of Q and A types in PlotQA.}}
\label{template_ans_stats}
\end{center}
\end{table}

\subsection{Sample Question Collection by Crowd-sourcing}
As the source data of PlotQA dataset is significantly richer in comparison to FigureQA and DVQA, we found it necessary to ask a larger set of annotators to create questions over these plots. 
However, creating questions for all the plots in our dataset would have been prohibitively expensive. 
We sampled $1,400$ plots across different types and asked workers on Amazon Mechanical Turk to create questions for these plots. 
We showed each plot to 5 different workers resulting in a total of $7,000$ questions. 
We specifically instructed the workers to ask complex reasoning questions which involved reference to multiple plot elements in the plots. 
We paid the workers USD $0.1$ for each question.

\subsection{Question Template Extraction \& Instantiation}

We manually analyzed the questions collected by crowdsourcing and divided them into a total of 74 templates. 
These templates were divided into 3 question categories. 
These question categories along with a few sample templates are shown below. See Table \ref{template_ans_stats} for statistics of different question and answer types in our dataset (please refer to the Supplementary material for further details).

\noindent\textbf{Structural Understanding}: 
    These are questions about the overall structure of the plot and do not require any quantitative reasoning. 
    Examples: ``How many different coloured bars are there?'', ``How are the legend labels stacked?''.

\noindent\textbf{Data Retrieval}:
    These questions seek data item for a single element in the plot.
    Examples: ``What is the number of tax payers in Myanmar in 2015?''.

\noindent\textbf{Reasoning}: 
    These questions either require numeric reasoning over multiple plot elements or a comparative analysis of different elements of the plot, or a combination of both to answer the question. 
    Examples: 
    ``In which country is the number of threatened bird species minimum?'', 
    ``What is the median banana production?'', 


We abstracted the questions into templates such as ``In how many $<$plural form of X\_label$>$, is the $<$Y\_label$>$ of/in $<$legend\_label$>$ greater than the average $<$Y\_label$>$ of/in $<$legend\_label$>$ taken over all $<$plural form of X\_label$>$?''. We could then generate multiple questions for each template by replacing X\_label, Y\_label, legend\_label, etc. by indicator variables, years, cities etc. from our curated data. However, this was a tedious task requiring a lot of manual intervention. For example, consider the indicator variable ``Race of students'' in Figure \ref{fig:sample_plots_PlotQA}. 
If we substitute this indicator variable as it is in the above template, it would result in a question, ``In how many cities, is the race of the students(\%) of Asian greater than the average race of the students (\%) of Asian taken over all cities?'', which sounds unnatural. To avoid this, we asked in-house annotators to carefully paraphrase these indicator variables and question templates. 
The paraphrased version of the above example was ``In how many cities, is the percentage of Asian students greater than the average percentage of Asian students taken over all cities?''. Such paraphrasing for every question template and indicator variable required significant manual effort. Using this semi-automated process we generated a total of \revision{$28,952,641$} questions. 
This approach of creating questions on real-world plot data with carefully curated question templates followed by manual paraphrasing is a key contribution of our work. 
The resultant PlotQA dataset is much closer to the real-world challenge of reasoning over plots, significantly improving on existing datasets.
\MK{Table \ref{tab:comparison_data_stats} summarizes the differences between PlotQA and these existing datasets such as FigureQA and DVQA. Note that (a) the number of unique answers in PlotQA is very large, (b) the questions in PlotQA are much longer, and (c) the vocabulary of PlotQA is more realistic than FigureQA or DVQA.}  


\section{Proposed Model}
\label{sec:proposed_model}

\MK{Existing models for VQA are of two types: (i) read the answer from the image (as in LoRRA) or (ii) pick the answer from a fixed vocabulary (as in SAN and BAN).} Such models work well for datasets such as DVQA where indeed all answers come from a fixed vocabulary (global or plot specific) but are not suited for PlotQA with a large number of OOV questions. 
\MK{Answering such questions involves various sub-tasks: (i) detect all the elements in the plot (bars, legend names, tick labels, etc), (ii) read the values of these elements, (iii) establish relations between the plot elements, e.g., creating tuples of the form \{country=Angola, year=2006, price of diesel = 0.4 \}, and (iv) reason over this structured data. Expecting a single end-to-end model to be able to do all of this is unreasonable.}
Hence, we propose a multi-staged pipeline to address each of the sub-tasks.

\begin{figure*}
\begin{center}
\includegraphics[width=\textwidth]{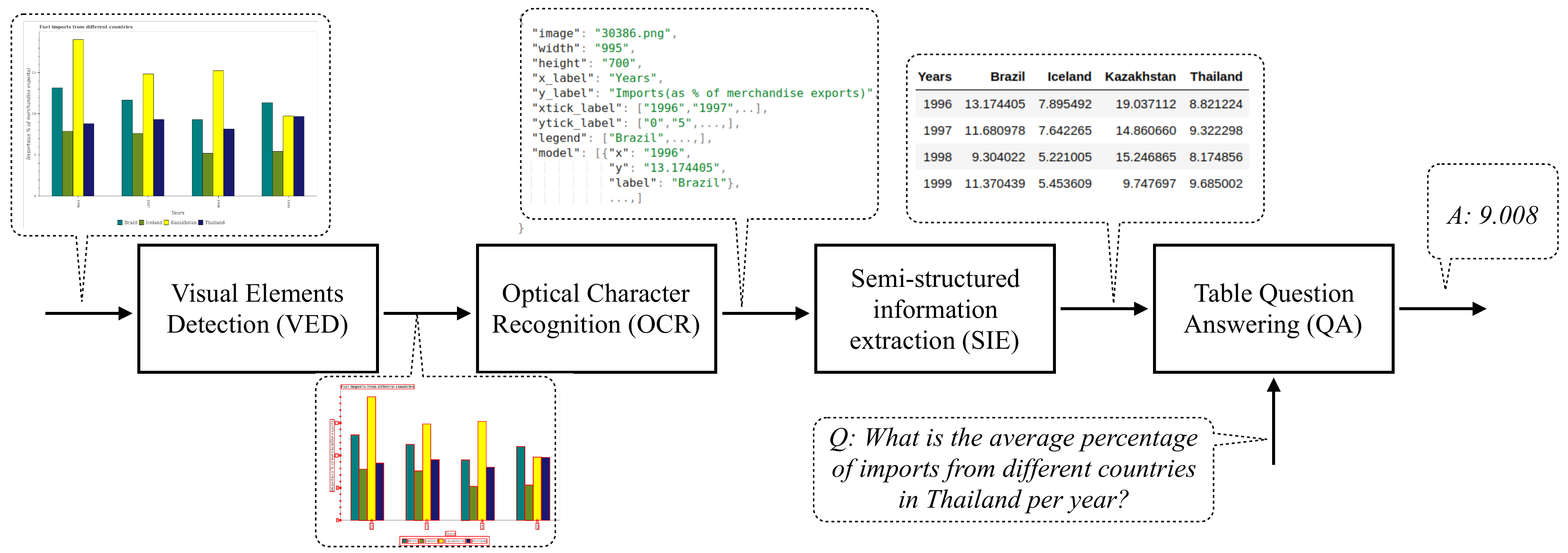}
\end{center}
\caption{Our proposed multi-staged modular pipeline for QA on scientific plots.}
\label{fig:pipeline}
\end{figure*}

\MK{We further note that for simpler questions which do not require reasoning and can be answered from a small fixed size vocabulary, such an elaborate pipeline is an overkill. As an illustration consider the question ``How many bars are there in the image?''. This does not require reasoning and can be answered based on visual properties of the image. For such questions, we have a simpler \textit{QA-as-classification} pipeline. As shown in Figure \ref{fig:pipeline}, our overall model is thus a hybrid model containing the following elements: (i) a binary classifier for deciding whether the given question can be answered from a small fixed vocabulary or needs more complex reasoning, and (ii) a simpler \textit{QA-as-classification} model to answer questions of the former type, and (iii) a multi-staged model containing four components as described below to deal with complex reasoning questions.}

\subsection{Visual Elements Detection (VED)}
The data bearing elements of a plot are of 10 distinct classes: the title, the labels of the $x$ and $y$ axes, the tick labels or categories (e.g., countries) on the $x$ and $y$ axis, the data markers in the legend box, the legend names, and finally the bars and lines in the graph.
Following existing literature (\cite{ClicheRMY17}, \cite{DVQA}), we refer to these elements as the {\em visual elements} of the graph. 
The first task is to extract all these visual elements by drawing bounding boxes around them and classifying them into the appropriate class. To this end, we can either apply object detection models such as RCNN, Fast-RCNN \cite{FastRCNN}, YOLO \cite{YOLO}, SSD \cite{LiuAESRFB16}, \textit{etc.} or instance segmentation models such as Mask-RCNN \cite{MaskRCNN}. \revision{Upon comparing all methods, we found that Faster R-CNN \cite{DBLP:conf/nips/RenHGS15} model along with Feature Pyramid Network(FPN) \cite{DBLP:conf/cvpr/LinDGHHB17} performed the best and hence we used it as our VED module.} 


\begin{table}[]
\footnotesize
\begin{center}
\begin{tabular}{|l|r|r|}
\hline
\textbf{Dataset Split} & \textbf{\#Images} & \textbf{\#QA pairs} \\ \hline
Train & 157,070 & 20,249,479 \\
Validation & 33,650 & 4,360,648 \\
Test & 33,657 & 4,342,514 \\ \hline
\textbf{Total} & 224,377 & 28,952,641 \\ \hline
\end{tabular}
\caption{\revision{PlotQA Dataset Statistics}}
\label{basic-stats-old}
\end{center}
\end{table}

\subsection{Object Character Recognition (OCR)}
Some of the visual elements such as title, legends, tick labels, \textit{etc.} contain numeric and textual data. For extracting this data from within these bounding boxes, we use a state-of-the-art OCR model \cite{Smith07}. We crop the detected visual element to its bounding box, convert the cropped textual image into gray-scale, resize and deskew it, and then pass it to an OCR module. 
Existing OCR modules perform well for machine-written English text, and indeed we found that a pre-trained OCR module\footnote{https://github.com/tesseract-ocr/tesseract} works well on our dataset. 

\subsection{Semi-Structured Information Extraction (SIE)}
The next stage of extracting the data into a semi-structured table is best explained with an example shown in Figure \ref{fig:pipeline}.
The desired output of SIE is shown in the table where the rows correspond to the ticks on the $x$-axis (1996, 1997, 1998, 1999), the columns correspond to the different elements listed in the legend (Brazil, Iceland, Kazakhstan, Thailand) and the $i$,$j$-th cell contains the value corresponding to the $x$-th tick and the $y$-th legend. 
The values of the $x$-tick labels and the legend names are available from the OCR module. 
The mapping of legend name to legend marker or color is done by associating a legend name to the marker or color whose bounding box is closest to the bounding box of the legend name.
Similarly, we associate each tick label to the tick marker whose bounding box is closest to the bounding box of the tick label. 
For example, we associate the legend name Brazil to the color ``Dark Cyan'' and the tick label 1996 to the corresponding tick mark on the $x$-axis. 
With this we have the 4 row headers and 4 column headers, respectively.
To fill in the 16 values in the table, there are again two smaller steps. 
First we associate each of the 16 bounding boxes of the 16 bars to their corresponding $x$-ticks and legend names. 
A bar is associated with an $x$-tick label whose bounding box is closest to the bounding box of the bar. 
To associate a bar to a legend name, we find the dominant color in the bounding box of the bar and match it with a legend name corresponding to that color.
Second, we need to find the value represented by each bar. 
We extract the height of the bar using bounding box information from the VED module and then search for the $y$-tick labels immediately above and below that height. 
We then interpolate the value of the bar based on the values of these bounding ticks.
With this we have the 16 values in the cells and thus have extracted all the information from the plot into a semi-structured table. The output of each stage  is discussed in the supplementary material.

\subsection{Table Question Answering (QA)}
The final stage of the pipeline is to answer questions on the semi-structured table. 
As this is similar to answering questions from the WikiTableQuestions dataset \cite{PasupatL15}, we adopt the same methodology as proposed in \cite{PasupatL15}. 
In this method, the table is converted to a knowledge graph and the question is converted to a set of candidate logical forms by applying compositional semantic parsing.
These logical forms are then ranked using a log-linear model and the highest ranking logical form is applied to the knowledge graph to get the answer. 
Note that with this approach the output is computed by a logical form that operates on the numerical data. 
This supports complex reasoning questions and also avoids the limitation of using a small answer vocabulary for multi-class classification as is done in existing work on VQA.
\textcolor{black}{There are recent neural approaches for answering questions over semi-structured tables such as \cite{NeelakantanLAMA16,HaugGG18}. Individually these models do not outperform the relatively simpler model of \cite{PasupatL15}, but as an ensemble they show a small improvement of only (1-2\%).}
To the best of our knowledge, there is only one neural method \cite{krishnamurthy-etal-2017-neural} which outperforms \cite{PasupatL15}, but the code for this model is not available which makes it hard to reproduce the results.

\begin{figure*}
\begin{center}
\includegraphics[width=\textwidth]{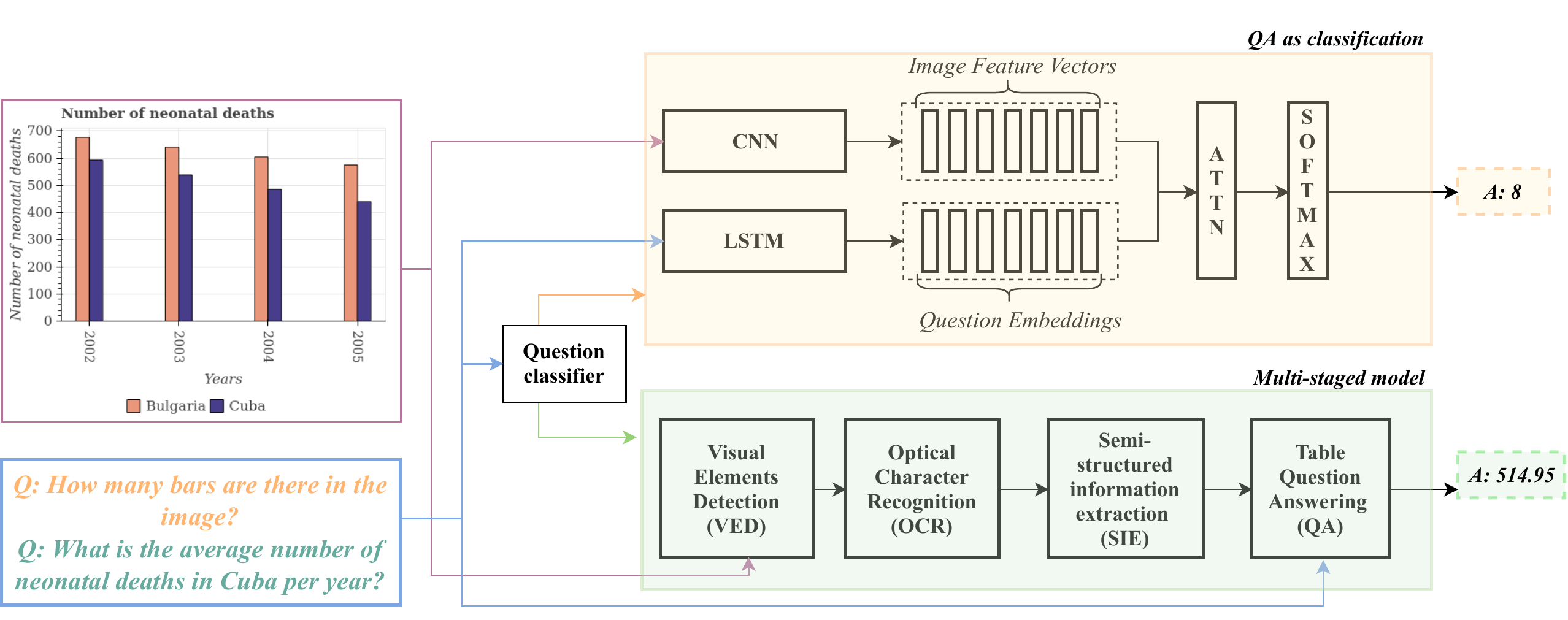}
\end{center}
\caption{Our proposed model containing (i) a question classifier for deciding whether the question can be answered from a fixed vocabulary (orange) or needs more complex reasoning (green), (ii) \textit{QA-as-classification} model to answer questions of the former type, and (iii) \textit{multi-staged model} as a pipeline of perception and QA modules for answering complex questions.}
\label{fig:our_model}
\end{figure*} 



\section{Experiments}
\label{sec:experiments}

\subsection{Train-Valid-Test Splits} 
By using different combinations of $841$ indicator variables and $160$ entities (years, countries, \textit{etc}), we created a total of $224,377$ plots. 
Depending on the context and type of the plot, we instantiated the $74$ templates to create meaningful \{question, answer\} pairs for each of the plots. 
We created train (70\%), valid (15\%) and test (15\%) splits (Table \ref{basic-stats-old}). 
\PN{The dataset and the crowd-sourced questions can be downloaded from the link: \url{bit.ly/PlotQA}}

\subsection{Models Compared}
We compare the performance of the following models:\\
\noindent - \textit{\textbf{IMG-only}}: This is a simple baseline where we just pass the image through a VGG19 and use the embedding of the image to predict the answer from a fixed vocabulary.\\
\noindent - \textit{\textbf{QUES-only}}: This is a simple baseline where we just pass the question through a LSTM and use the embedding of the question to predict the answer from a fixed vocabulary.\\ 
\noindent - \textit{\textbf{SAN}} \cite{SAN}: This is an encoder-decoder model with a multi-layer stacked attention \cite{Attention} mechanism. It obtains a representation for the image using a deep CNN and a representation for the query using LSTM. It then uses the query representation to locate relevant regions in the image and uses this to pick an answer from a fixed vocabulary.\\
\noindent - \textit{\textbf{SANDY}} \cite{DVQA}: This is the best performing model on the DVQA dataset and is a variant of SAN. Unfortunately, the code for this model is not available and the description in the paper was not detailed enough for us to reimplement it.\footnote{We have contacted the authors and while they are helpful in sharing various details, they do not have access to the original code now.} Hence, we report the numbers for this model only on DVQA (from the original paper).\\
\MK{\noindent - \textit{\textbf{LoRRA}} \cite{singh2019TowardsVM}: This is the recently proposed model on the TextVQA dataset. It concatenates the image features extracted from pre-trained ResNet-152 \cite{DBLP:conf/cvpr/HeZRS16} model with the region based features extracted from Faster-RCNN \cite{Detectron2018} model. It then reads the text present in the image using a pre-trained OCR module and incorporates an attention mechanism to reason about the image and the text. Finally, it does multi-class classification where the answer either comes from a fixed vocabulary or is copied from the text in the image.
}\\
\MK{\noindent - \textit{\textbf{BAN}} \cite{DBLP:conf/nips/KimJZ18}: This model exploits bilinear interactions between two groups of input channels, \textit{i.e.}, between every question word (GRU \cite{DBLP:conf/emnlp/ChoMGBBSB14} features) and every image region (pre-trained Faster-RCNN \cite{DBLP:conf/nips/RenHGS15} object features). It then uses low-rank bilinear pooling \cite{DBLP:conf/iclr/KimOLKHZ17} to extract the joint distribution for each pair of channels. BAN accumulates 8 such bilinear attention maps which are then fed to a two-layer perceptron classifier to get the final joint distribution over answers from a fixed vocabulary.}\\
\noindent - \textit{\textbf{Our Model}}:  This proposed model shown in Figure~\ref{fig:our_model} with two model paths.
The training data for the binary classification is generated by comparing the performance of the individual models: For a given question, the label is set to 1 if the performance of QA-as-classification model is better than the multi-stage pipeline, and 0 otherwise. 
       We use an LSTM to represent the input question and then perform binary classification on this representation.

\subsection{Training Details}


 
\noindent\textbf{SAN:} We used an existing implementation of SAN\footnote{https://github.com/TingAnChien/san-vqa-tensorflow} for the initial baseline results. Image features are extracted from the last pooling layer of VGG19 network. Question features are the last hidden state of the LSTM. Both the LSTM hidden state and 512-d image feature vector at each location are transferred to a 1024-d vector by a fully connected layer, and added and passed through a non-linearity (tanh). The model was trained using Adam \cite{Adam} with an initial learning rate of $0.0003$ and a batch size of 128 for 25,000 iterations.

\noindent\textbf{Our model:}
The binary question classifier in the proposed model contains a 50-dimensional word embedding layer followed by an LSTM with 128 hidden units. The output of the LSTM is projected to 256 dimensions and this is then fed to the output layer. The model is trained for 10 epochs using RMSProp with an initial learning rate of 0.001. Accuracy on the validation set is $87.3\%$.
Of the 4 stages of the multi-stage pipeline, only two require training, \textit{viz.}, Visual Elements Detection (VED) and Table Question Answering (QA). 
As mentioned earlier, for VED we train a \revision{variant of Faster R-CNN \cite{DBLP:conf/cvpr/LinDGHHB17} with FPN} using the bounding box annotations available in PlotQA. 
We trained the model with a batch size of 32 for $200,000$ steps. We used RMSProp with an initial learning rate of $0.004$. 
For Table QA, we trained the model proposed in \cite{PasupatL15} using questions from our dataset and the corresponding \textcolor{black}{ground truth tables}. 


\subsection{Evaluation Metric} We used accuracy as the evaluation metric. Specifically, for textual answers (such as India, CO2, etc.) the model's output was considered to be correct only if the predicted answer exactly matches the true answer. However, for numeric answers with floating point values, an exact match is a very strict metric 
We relax the measure to consider an answer to be correct as if it is within 5\% of the correct answer. 



\subsection{Human Accuracy on PlotQA dataset}
\revision{To assess the difficulty of the PlotQA dataset, we report  human accuracy on a small subset of the Test split of the dataset. With the help of in-house annotators, we were able to evaluate $5,860$ questions grounded in $160$ images. Human accuracy on this subset is found to be $80.47\%$. We used the evaluation metric as defined in section 5.4. Most human errors were due to numerical precision as it is difficult to find the exact value from the plot even with a 5\% margin.}


\section{Observations and Results}

\noindent \textbf{1. Evaluating models on PlotQA dataset (Table \ref{plotqa-overall}):}
\textcolor{black}{The baselines IMG-only and QUES-only performed poorly with an accuracy of \revision{$4.84\%$ and $5.35\%$} respectively.}
Existing models (SAN, BAN, LoRRA) perform poorly on this dataset. In particular, BAN and LoRRA have an abysmal accuracy of less than 1\%. This is not surprising given that both models are not designed to answer OOV questions. Further, the original VQA tasks for which BAN was proposed does not have any complex numerical reasoning questions as found in PlotQA. Similarly, LoRRA was designed only for text based answers and not for questions  requiring numeric reasoning. Note that we have used the original code \cite{singh2018pythia} released by the authors of these models. Given the specific focus and limited capabilities of these existing models it may even seem unfair to evaluate these models on our dataset but we still do so for the sake of completeness and to highlight the need for better models. Lastly, our model gives the best performance of 22.52\% on the PlotQA dataset.\\
\PN{\textbf{Ablation Study of proposed method}: Table \ref{ablation} presents the details of the ablation study of the proposed method for each question type (structural, data retrieval, reasoning) and each answer type (binary, fixed vocabulary, OOV). QA-as-classification performs very well on Yes/No questions and moderately well on Fixed vocab. questions with a good baseline aggregate accuracy of 7.76\%. It performs poorly on Open vocab. question, failing to answer almost all the 3,507,045 questions in this category. On the other hand, the QA-as-multi-stage pipeline fails to answer correctly any of the Yes/No questions, performs moderately well on Fixed vocab. questions, and answers correctly some of the hard Open vocab. questions. Our model combines the complementary strengths of QA-as-classification and QA-as-multi-stage pipeline achieving the highest accuracy of 22.52\%. In particular, the performance improves significantly for all Fixed Vocab. questions, while retaining the high accuracy of QA-as-classification on Yes/No questions and QA-as-multi-stage pipeline's performance on Open vocab. We acknowledge that the accuracy is significantly lower than human performance. This establishes that the dataset is challenging and raises open questions on models for visual reasoning.}

\begin{figure}[h!]
\begin{center}
\includegraphics[scale=0.3]{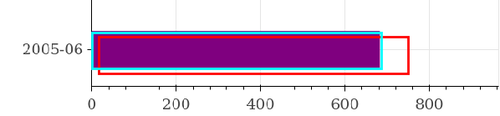}
\caption{Ground-truth (cyan) and predicted (red) boxes.}
\label{VED}
\end{center}
\end{figure}

\noindent \textbf{2. Analysis of the pipeline}
We analyze the performance of VED, OCR and SIE modules in the pipeline. \\
\noindent \textbf{VED:} Table \ref{maskrcnn_ved} shows that the VED module performs reasonably well at an Intersection Over Union (IOU) of 0.5.
For higher IOUs of 0.75 and 0.9, the accuracy falls drastically. For instance, at IOU of 0.9, dotlines are detected with an accuracy of under 20\%.
Clearly, such inaccuracies would lead to incorrect table generation and subsequent QA.
This brings out an interesting difference between this task and other instance segmentation tasks where the margin of error is higher (where IOU of 0.5 is accepted).
A small error in visual element detection as indicated by mAP scores of 75\% is considered negligible for VQA tasks, however for PlotQA small errors can cause significantly misaligned table generation and subsequent QA.
\textcolor{black}{We illustrate this with an example given in Figure \ref{VED}. The predicted red box having an IOU of 0.58 estimates the bar size as 760 as opposed to ground truth of 680, significantly impacting downstream QA accuracy.} 
\\
\noindent \textbf{OCR:} We evaluate the OCR module in standalone/oracle mode and pipeline mode in Table \ref{ocr-acc}.  In the oracle mode, we feed ground truth boxes to the OCR model whereas in the pipeline model we perform OCR on the output of the VED module. We observe only a small drop in performance from 97.06\% (oracle) to 93.10\% (after VED), which indicates that the OCR module is robust to the reduction in VED module's accuracy at higher IOU as it does not depend on the class label or the exact position of bounding boxes. \\
\MK{ 
\noindent \textbf{SIE:} We now evaluate the performance of the SIE module. We consider each cell in the table to be a tuple of the form \{row header, column header, value \} (e.g., \{Poland, 1964, 10000 tractors). We consider all the tuples extracted by the SIE module with the tuples present in the ground truth table to compute the F1-score. Even though Table \ref{maskrcnn_ved} suggests that the VED model is very accurate with a mAP@0.5 of 96.43\%, we observe that the F1-score for table extraction is only 0.68. This indicates that many values are not being extracted accurately due to the kind of errors shown in Figure \ref{VED} where the bounding box has a high overlap with the true box. We thus need better plot VED modules which can predict tighter bounding boxes (higher mAP at IOU of 0.9) around the plot's visual and textual elements. Inaccurate VED module leads to erroneous tables which further affects the downstream QA accuracy.}\\
- In summary, a highly accurate VED for structured images 
is an open challenge to improve reasoning over plots. 

\noindent \textbf{3. Evaluating new models on the existing DVQA dataset (Table \ref{dvqa-overall}):} 
The proposed model performs better than the existing models (SAN and SANDY-OCR) establishing a new SOTA result on DVQA. 
The higher performance of the proposed hybrid model in comparison to SAN (in contrast to the PlotQA results) suggests that the extraction of the structured table is more accurate on the DVQA dataset.
This is because of the limited variability in the axis and tick labels and shorter length (one word only) of labels. 

\begin{table}[t]
\footnotesize
\begin{center}
\begin{tabular}{|l|r|r|}
\hline
\multicolumn{1}{|c|}{\textbf{Model}} & \multicolumn{1}{c|}{\begin{tabular}[c]{@{}c@{}}\textbf{DVQA} \textbf{(TEST)}\end{tabular}} & \multicolumn{1}{c|}{\begin{tabular}[c]{@{}c@{}}\textbf{DVQA} \textbf{(TEST-NOVEL)}\end{tabular}} \\ \hline
SAN & 32.1\% & 30.98\% \\
SANDY-OCR & 45.77\% & 45.81\% \\
Our Model & \textbf{57.99\%} & \textbf{59.54\%} \\ \hline
\end{tabular}
\caption{Accuracy of different models on DVQA dataset.}
\label{dvqa-overall}
\end{center}
\end{table}


\begin{table}[t]
\footnotesize
\begin{center}
\begin{tabular}{|l|c|c|c|c|c|c|}
\hline
\textbf{Models} & \textbf{IMG} & \textbf{QUES} & \textbf{BAN} & \textbf{LoRRA} & \textbf{SAN} & \textbf{\begin{tabular}[c]{@{}c@{}}Our\\ Model\end{tabular}} \\ \hline
\multicolumn{1}{|c|}{\begin{tabular}[c]{@{}c@{}}\textbf{Accuracy}\end{tabular}} & \multicolumn{1}{r|}{4.84} & \multicolumn{1}{r|}{5.35} & \multicolumn{1}{r|}{0.01} & \multicolumn{1}{r|}{0.02} & \multicolumn{1}{r|}{7.76} & \multicolumn{1}{r|}{\textbf{22.52}} \\ \hline
\end{tabular}
\caption{\revision{Accuracy (in \%) of different models on PlotQA.}}
\label{plotqa-overall}
\end{center}
\end{table}

\begin{table}[t]
\footnotesize
\begin{center}
\setlength\tabcolsep{2.5pt}
\begin{tabular}{|l|l|rrr|}
\hline
\multicolumn{5}{|c}{\textbf{Accuracy (in \%)}} \vline \\
\hline
\parbox{1.3cm}{\textbf{Model}\\\textbf{(Agg. acc.)}} & \parbox{1.3cm}{\textbf{Q Type\textbackslash} \\\textbf{A type}} & \textbf{Structural} & \textbf{Data Retrieval} & \textbf{Reasoning}  \\ \hline
\multirow{3}{*}{\parbox{1.3cm}{\PN{Human\\ Baseline} \\ (80.47)}}
& Yes/No  & 99.77 & 100 & 76.51 \\
& Fixed vocab. & 99.29 & 83.31 & 59.97\\
& Open vocab. & NA & 87.58 & 58.01 \\ \hline
\multirow{3}{*}{\parbox{1.3cm}{QA as classification \\ (7.76)}} 
& Yes/No  & 91.12 & 97.32  & 62.75 \\
& Fixed vocab. & 66.85 & 30.76 & 16.03  \\
& Open vocab. & NA & 0.00 & 0.00 \\ \hline
\multirow{3}{*}{\parbox{1.3cm}{\PN{Multistage Pipeline} \\ (18.46)}}
& Yes/No  & 0.00 & 0.00 & 0.00 \\
& Fixed vocab. & 42.12 & 16.07 & 7.24  \\
& Open vocab. & NA & 57.39 & 14.95  \\ \hline
\multirow{3}{*}{\parbox{1.3cm}{Our Model \\ \textbf{(22.52)}}}
& Yes/No  & \textbf{91.12} & \textbf{97.32} & \textbf{62.75} \\
& Fixed vocab. & \textbf{66.86} & \textbf{22.64} & \textbf{7.95}  \\
& Open vocab. & NA & \textbf{57.39} & \textbf{14.95} \\ \hline
\end{tabular}
\caption{\PN{Ablation study of proposed method on PlotQA. Note that there are no open vocab. answers for \textit{Structural Understanding} question templates (see Table \ref{sample_qs_stats}).}}
\label{ablation}
\end{center}
\end{table}


\begin{table}[t]
\footnotesize
\begin{center}
\begin{tabular}{|l|r|r|r|}
\hline 
\textbf{Class} & \textbf{AP@0.5} & \textbf{AP@0.75} & \textbf{AP@0.9} \\ \hline
Title & 100.00\% & 78.83\% & 0.22\% \\
Bar & 95.84\% & 94.30\% & 85.54\% \\
Line & 72.25\% & 62.04\% & 37.65\% \\
Dotline & 96.30\% & 95.14\% & 18.07\% \\
X-axis Label & 99.99\% & 99.99\% & 99.09\% \\
Y-axis Label & 99.90\% & 99.90\% & 99.46\% \\
X-tick Label & 99.92\% & 99.74\% & 96.04\% \\
Y-tick Label & 99.99\% & 99.97\% & 96.80\% \\
Legend Label & 99.99\% & 99.96\% & 93.68\% \\
Legend Preview & 99.95\% & 99.94\% & 96.30\% \\ \hline
\textbf{mAP} & \textbf{96.43\%} & \textbf{92.98\%} & \textbf{72.29\%}\\ \hline
\end{tabular}
\caption{\MK{VED Module's Accuracy on PlotQA dataset}}
\label{maskrcnn_ved}
\end{center}
\end{table}

\begin{table}[t]
\footnotesize
\begin{center}
\begin{tabular}{|l|r|r|}
\hline
& \textbf{Oracle} & \textbf{After VED}\\ \hline
Title & 99.31\% & 94.6\% \\
X-axis Label & 99.94\% & 95.5\% \\
Y-axis Label & 98.43\% & 97.07\% \\
X-tick Label & 94.8\% & 91.38\% \\
Y-tick Label & 93.38\% & 88.07\% \\
Legend Label & 98.53\% & 91.99\% \\
\hline
\textbf{Total} & \textbf{97.06\%} &\textbf{ 93.10\%} \\
\hline
\end{tabular}
\caption{\PN{OCR Module Accuracy on the PlotQA dataset.}}
\label{ocr-acc}
\end{center}
\end{table}

\section{Conclusion}
We introduce the PlotQA dataset to reduce the gap between existing synthetic plot datasets and real-world plots and question templates.
Analysis of existing VQA models on PlotQA reveals that they perform poorly for Open Vocabulary questions. This is not surprising as these models were not designed to handle complex questions which require numeric reasoning and OOV answers. We propose a hybrid model with separate pipelines for handling (i) simpler questions which can be answered from a fixed vocabulary and (ii) complex questions with OOV answers. For OOV questions, we propose a pipelined approach that combines visual element detection and OCR with QA over tables. The proposed model gives state-of-the-art results on both the DVQA and PlotQA datasets. 
Further analysis of our pipeline reveals the need for more accurate visual element detection to improve reasoning over plots.

{\small
\bibliographystyle{ieee}
\bibliography{main}
}

\newpage

\section*{\textcolor{white}{Pass}}
\newpage
\section*{Appendices}
The supplementary material is organised in the following manner:
Section \ref{kg_creation} describes the methodology of knowledge graph creation from structured data.
In Section \ref{failure_cases}, we further analyse our proposed pipeline and discuss the errors in each stage.
In Section \ref{plot_samples}, we provide sample plots from the PlotQA dataset.
In Section \ref{templates}, we list all the 74 question templates that were formulated from the crowd sourced questions.
\section{Construction of knowledge graph from structured data} \label{kg_creation} 
\begin{figure}[h]
\centering
\begin{subfigure}{.45\textwidth}
\centering
\includegraphics[scale=0.4]{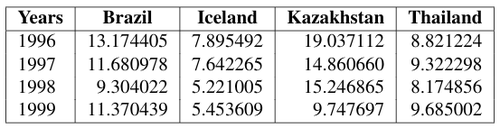}
\caption{\PN{Semi-Structured Table}}
\label{fig:sie_table}
\end{subfigure}
\begin{subfigure}{.45\textwidth}
\centering
\includegraphics[scale=0.55]{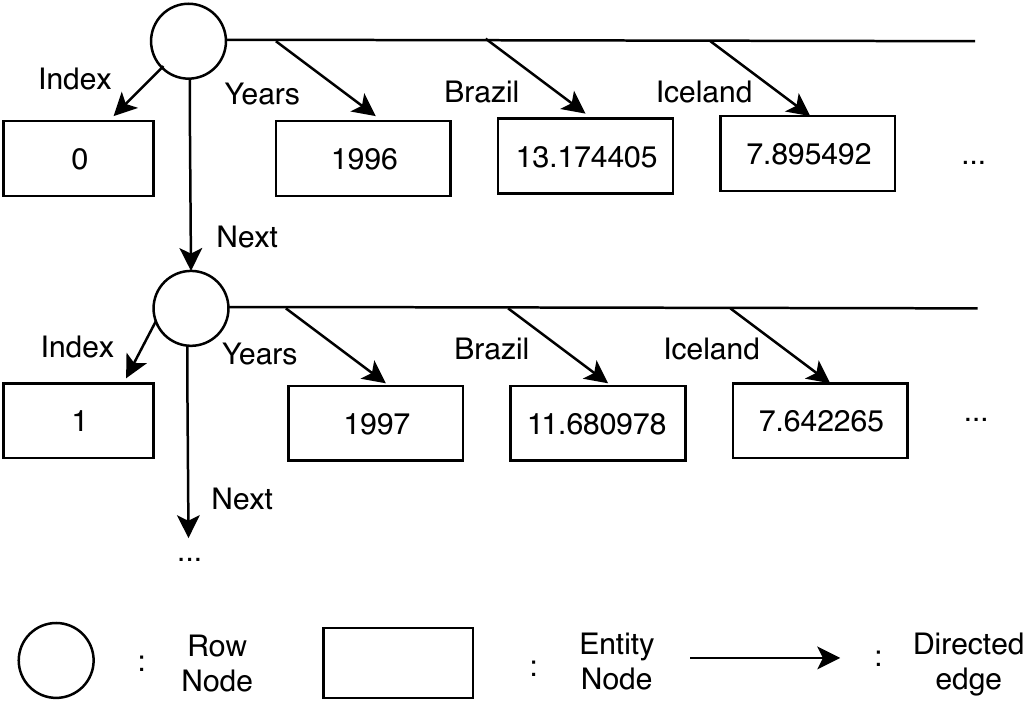}
\caption{\PN{Knowledge Graph}}
\label{fig:flow}
\end{subfigure}
\caption{\PN{An example of the knowledge graph constructed from the semi-structured table.}}
\label{KG}
\end{figure}

Following \cite{PasupatL15}, we convert the semi-structured table into a knowledge graph which has two types of nodes \textit{viz.} row nodes and entity nodes. The rows of the table become row nodes, whereas the cells of each row become the entity nodes in the graph. Directed edges exist from the row nodes to the entity nodes of that column and the corresponding table column header act as edge-labels. An example of knowledge graph of the semi-structured table given in Figure \ref{fig:sie_table} is shown in Figure \ref{fig:flow}. For reasoning on the knowledge graph, we adopted the same methodology as given in \cite{PasupatL15}. The questions are converted to a set of candidate logical forms by applying compositional semantic parsing. Each of these logical forms is then ranked using a log-linear model and the highest ranking logical form is applied to the knowledge graph to get the final answer. 


\section{\PN{Some failure cases}}
\label{failure_cases}

In this section, we visualize the output of each stage in our multistage pipeline and show some interesting failure cases with the help of an example.

\begin{figure*}
\centering
\begin{subfigure}{.45\textwidth}
\centering
\includegraphics[scale=0.4]{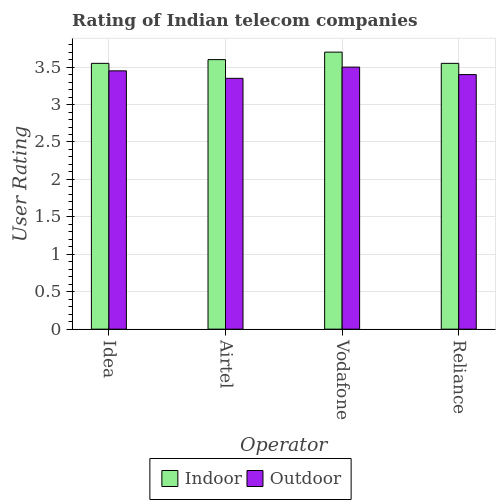}
\caption{Input plot image}
\end{subfigure}
\hspace{0.5cm}
\begin{subfigure}{.45\textwidth}
\centering
\includegraphics[scale=0.4]{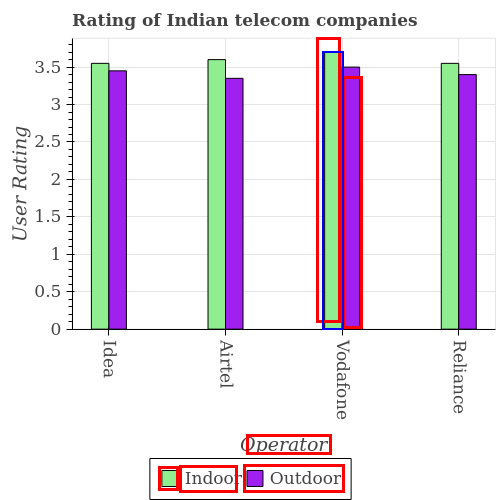}
\caption{Few examples of the predicted bounding boxes}
\end{subfigure}
\caption{\PN{Errors made by the VED stage (highlighted in red). }}
\label{fig:error_analysis_ved_1}
\end{figure*}

\begin{figure*}
\centering
\begin{subfigure}{.45\textwidth}
\centering
\includegraphics[scale=0.35]{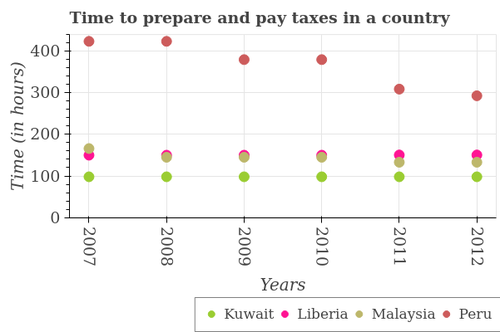}
\caption{Input plot image}
\end{subfigure}
\hspace{0.5cm}
\begin{subfigure}{.45\textwidth}
\centering
\includegraphics[scale=0.35]{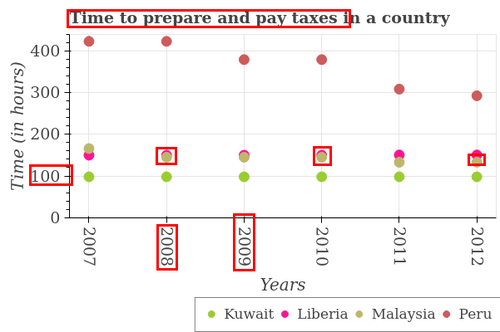}
\caption{Few examples of the predicted bounding boxes}
\end{subfigure}
\caption{\PN{Errors made by the VED stage (highlighted in red).}}
\label{fig:error_analysis_ved_2}
\end{figure*}

\begin{figure*}
\centering
\begin{subfigure}{.45\textwidth}
\centering
\includegraphics[scale=0.4]{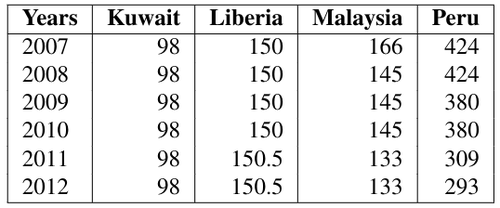}
\caption{Oracle table generated using ground-truth annotations}
\end{subfigure}
\hspace{0.5cm}
\begin{subfigure}{.45\textwidth}
\centering
\includegraphics[scale=0.4]{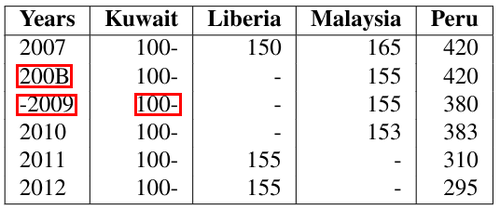}
\caption{Generated Semi-Structured Table}
\end{subfigure}
\caption{\PN{Errors made by the OCR and SIE stage (highlighted in red). Note that most of these errors have been propagated from the VED stage.}}
\label{fig:error_analysis_sie_2}
\end{figure*}

- \textbf{VED Stage:} Although the bounding boxes predicted by Faster R-CNN coupled with Feature Pyramid Network (FPN) fit reasonably well at an IOU of 0.5, it is not acceptable as the values extracted from these bounding boxes will lead to incorrect table generation and subsequent QA. Example: In Figure \ref{fig:error_analysis_ved_1}b, consider the bar representing the ``Indoor User Rating'' value for ``Vodafone''. The overlap between the ground-truth box (blue) and the predicted box (red) is higher than 0.5 but the values extracted from the detected box is $4.0$ as opposed to the actual value which is $3.73$. Another interesting failure case is shown in Figure \ref{fig:error_analysis_ved_2}b. There are multiple overlapping data points and the model is able to detect only one of the points. This leads to incomplete table generation as shown in Figure \ref{fig:error_analysis_sie_2}b where the values for Liberia for the years 2008, 2009 and 2010 could not be extracted. This small error might be acceptable for other VQA tasks but for PlotQA these small errors will escalate to multiple incorrect answers.

\begin{figure*}
\centering
\begin{subfigure}{.45\textwidth}
\centering
\includegraphics[scale=0.45]{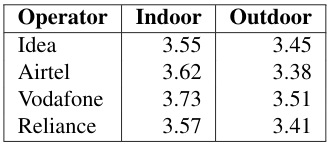}
\caption{Oracle table generated using ground-truth annotations}
\end{subfigure}
\hspace{0.5cm}
\begin{subfigure}{.45\textwidth}
\centering
\includegraphics[scale=0.45]{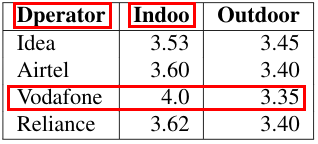}
\caption{Generated Semi-Structured Table}
\end{subfigure}
\caption{\PN{Errors made by the OCR and SIE stage (highlighted in red). Note that most of these errors have been propagated from the VED stage.}}
\label{fig:error_analysis_sie_1}
\end{figure*}

- \textbf{OCR stage:}
A slight misalignment in the bounding boxes predicted by VED module causes significant errors while extracting the text.
Example: In Figure \ref{fig:error_analysis_ved_1}b, consider the box enclosing the legend label ``Indoor''. The rightmost edge of the predicted bounding box is drawn over the letter ``r'', which makes the OCR module incorrectly recognize the text as ``Indoo''. A similar error is made while performing OCR on the X-axis title which is read as ``Dperator'' instead of ``Operator''. \PN{Consider another example where there are misaligned bounding boxes on axes tick-labels as shown in Figure \ref{fig:error_analysis_ved_2}b. The values extracted are 200B, -2009 and -100 as opposed to the ground-truth values 2008, 2009 and 100. This slight error leads to incorrect column name in the subsequent generated tables (Figure \ref{fig:error_analysis_sie_1}b and Figure \ref{fig:error_analysis_sie_2}b) and incorrect answers to all the questions pertaining to these labels as shown in Table \ref{tab:error_analysis_table_1} and Tale \ref{tab:error_analysis_table_2}.
}

- \textbf{SIE stage:} Figure \ref{fig:error_analysis_sie_1}a shows the oracle table which is generated by using the ground-truth annotations and Figure \ref{fig:error_analysis_sie_1}b shows the table generated after passing the plot image through the different stages of our proposed multistage pipeline. It is evident from the generated table that the errors propagated from the VED and the OCR stage has lead to an incorrect table generation.

- \textbf{QA stage:} In Table \ref{tab:error_analysis_table_1} and Table \ref{tab:error_analysis_table_2} we compare the answer predictions made by different models with the ground-truth answer on randomly sampled questions. Note that, our proposed model combines the complementary strengths of both, QA-as-classification and QA as multistage pipeline, models.

\begin{table*}[h!]
\centering
\begin{tabular}{|l|r|r|r|r|}
\hline
\multicolumn{1}{|c|}{\textbf{Question}} & \multicolumn{1}{c|}{\textbf{\begin{tabular}[c]{@{}c@{}}Ground \\ Truth\end{tabular}}} & \multicolumn{1}{c|}{\textbf{\begin{tabular}[c]{@{}c@{}}QA as\\ classification\end{tabular}}} & \multicolumn{1}{c|}{\textbf{\begin{tabular}[c]{@{}c@{}}Multistage \\ Pipeline\end{tabular}}} & \multicolumn{1}{c|}{\textbf{\begin{tabular}[c]{@{}c@{}}Our \\ Model\end{tabular}}} \\ \hline
\textbf{Q1.} What is the average indoor user rating per operator? & 3.62 & 500 & 3.54 & 3.54 \\ 
\textbf{Q2.} What is the total user rating for Vodafone in the graph? & 7.24 & 5 & 7.35 & 7.35 \\ 
\textbf{Q3.} What is the label or title of the X-axis? & Operator & Years & Dperator & Dperator \\ 
\textbf{Q4.} What is the indoor user rating of Airtel? & 3.62 & 0 & 3.60 & 3.60 \\ 
\textbf{Q5.} How many groups of bars are there? & 4 & 4 & 4 & 4 \\ 
\begin{tabular}[c]{@{}l@{}}\textbf{Q6.} What is the ratio of indoor user rating of Airtel to that\\ of the outdoor user rating of Vodafone?\end{tabular} & 1.03 & No & 3.35 & 3.35 \\ 
\begin{tabular}[c]{@{}l@{}}\textbf{Q7.} What is the difference between the highest and lowest\\ outdoor user rating?\end{tabular} & 0.13 & 1.5 & 0.10 & 0.10 \\ 
\begin{tabular}[c]{@{}l@{}}\textbf{Q8.} Does ``Indoor'' appear as one of the legend-labels in\\ the graph?\end{tabular} & Yes & Yes & 1.0 & Yes \\ 
\begin{tabular}[c]{@{}l@{}}\textbf{Q9.} For how many operators are the indoor user rating\\ greater than the average outdoor user rating taken \\ over all operators?\end{tabular} & 4 & 4 & 3.4 & 4 \\ 
\begin{tabular}[c]{@{}l@{}}\textbf{Q10.} Is the sum of outdoor user rating in Reliance and\\ Idea greater than the maximum outdoor user rating \\ across all operators?\end{tabular} & Yes & Yes & 6.85 & Yes \\ \hline
\end{tabular}
\caption{Answers predicted by different models on the sample questions.}
\label{tab:error_analysis_table_1}
\end{table*}

\begin{table*}[h!]
\centering
\begin{tabular}{|l|r|r|r|r|}
\hline
\multicolumn{1}{|c|}{\textbf{Question}} & \multicolumn{1}{c|}{\textbf{\begin{tabular}[c]{@{}c@{}}Ground \\ Truth\end{tabular}}} & \multicolumn{1}{c|}{\textbf{\begin{tabular}[c]{@{}c@{}}QA as\\ classification\end{tabular}}} & \multicolumn{1}{c|}{\textbf{\begin{tabular}[c]{@{}c@{}}Multistage \\ Pipeline\end{tabular}}} & \multicolumn{1}{c|}{\textbf{\begin{tabular}[c]{@{}c@{}}Our \\ Model\end{tabular}}} \\ \hline
\textbf{Q1.} How are the legend-labels stacked? & horizontal & horizontal & horizontal & horizontal \\
\begin{tabular}[c]{@{}l@{}}\textbf{Q2.} What is the time (in hours) required to\\ prepare and pay taxes in Kuwait in 2008?\end{tabular} & 98 & 100 & 100- & 100- \\
\begin{tabular}[c]{@{}l@{}}\textbf{Q3.} What is the difference between in time (in \\ hours) required to prepare and pay taxes in \\ Kuwait in 2009 and the time (in hours) required \\ to prepare and pay taxes in Liberia in 2008?\end{tabular} & -52 & -0.5 & 100- & 100- \\
\begin{tabular}[c]{@{}l@{}}\textbf{Q4.} What is the difference between the highest \\ and  the lowest time (in hours) required to \\ prepare and pay taxes in Peru?\end{tabular} & 131 & 500 & 125 & 125 \\
\begin{tabular}[c]{@{}l@{}}\textbf{Q5.} Is it the case that every year the sum of time \\ (in hours) required to prepare and pay taxes in \\ Peru and Kuwait is greater than the sum of the \\ time (in hours) required to prepare and pay \\ taxes in Liberia and Malaysia?\end{tabular} & Yes & Yes & 320 & Yes \\ \hline
\end{tabular}
\caption{Answers predicted by different models on the sample questions.}
\label{tab:error_analysis_table_2}
\end{table*}

\section{Samples from the PlotQA dataset}
\label{plot_samples}
\begin{figure*}
\centering
\begin{subfigure}{.45\textwidth}
\centering
\includegraphics[scale=0.35]{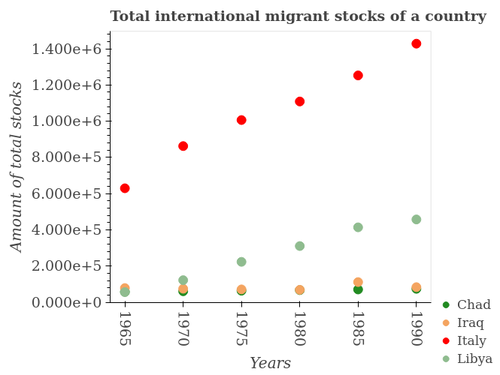}
\caption*{
\textbf{Q7}: What is the difference between two consecutive major ticks on the Y-axis?\\
\textbf{A}: $2.000\mathrm{e}+5$
}
\end{subfigure}
\hspace{0.5cm}
\begin{subfigure}{.45\textwidth}
\centering
\includegraphics[scale=0.35]{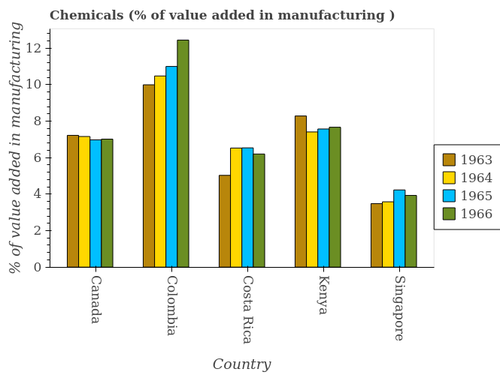}
\caption*{\textbf{Q8}: In how many cases, is the number of bars for a given year not equal to the number of legend labels?\\
\textbf{A}: 0
}
\end{subfigure}
\begin{subfigure}{.45\textwidth}
\centering
\includegraphics[scale=0.35]{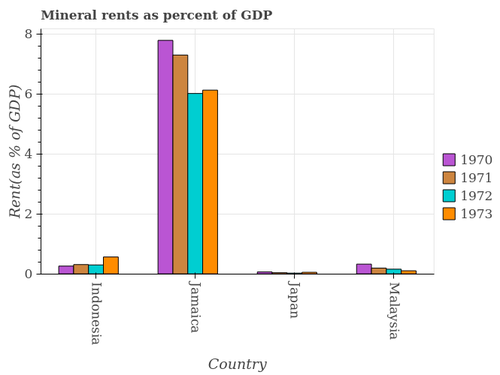}
\caption*{\textbf{Q9}: In how many countries, is the mineral rent (as \% of GDP) in 1970 greater than the average mineral rent (as \% of GDP) in 1970 taken over all countries?\\
\textbf{A}: 1
}
\end{subfigure}
\hspace{0.5cm}
\begin{subfigure}{.45\textwidth}
\centering
\includegraphics[scale=0.35]{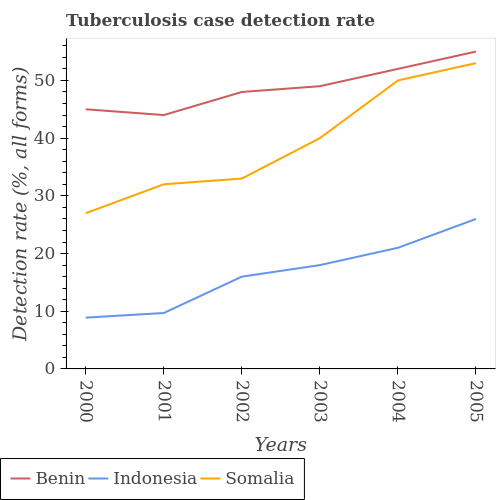}
\caption*{\textbf{Q10}: What is the total tuberculosis detection rate in Indonesia?\\
\textbf{A}: 101
}
\end{subfigure}
\begin{subfigure}{.45\textwidth}
\centering
\includegraphics[scale=0.32]{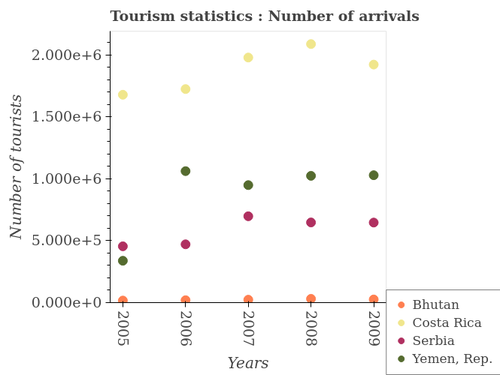}
\caption*{\textbf{Q11}: Is it the case that in every year, the sum of the number of tourists in Costa Rica and Serbia greater than the number of tourists in Bhutan?\\
\textbf{A}: Yes
}
\end{subfigure}
\hspace{0.5cm}
\begin{subfigure}{.45\textwidth}
\centering
\includegraphics[scale=0.42]{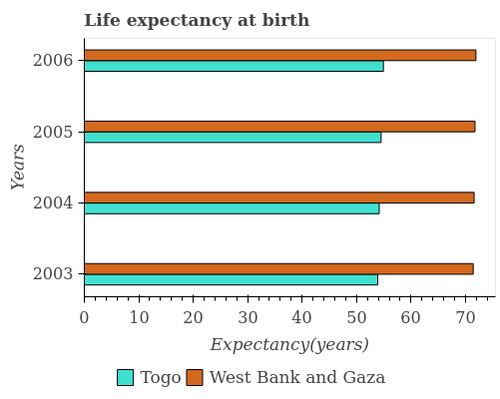}
\caption*{\textbf{Q12}: How many bars are there on the $2^{nd}$ tick from the top?\\
\textbf{A}: 2
}
\end{subfigure}
\caption{Sample \{plot, question, answer\} triplet present in the PlotQA dataset.}
\label{fig:supple_dataset_examples}
\end{figure*}

\PN{Few examples of the \{plot, question, answer\} triplets from the PlotQA dataset are shown in Figure \ref{fig:supple_dataset_examples}. For each of the plots, most of the question templates discussed in section \ref{templates} are applicable but depending on the context of the plot, their language varies from it's surface form.}
\begin{figure*}
\centering
\begin{subfigure}{.45\textwidth}
\centering
\includegraphics[scale=0.3]{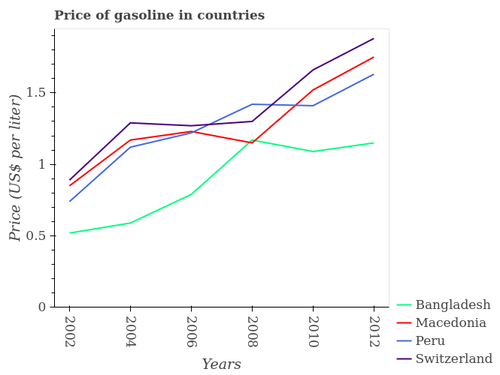}
\caption*{\textbf{Q1}: What is the difference between price of gasoline in Switzerland and price of gasoline in Macedonia in 2008?\\
\textbf{A}: 0.15}
\end{subfigure}
\hspace{0.2cm}
\begin{subfigure}{.45\textwidth}
\centering
\includegraphics[scale=0.35]{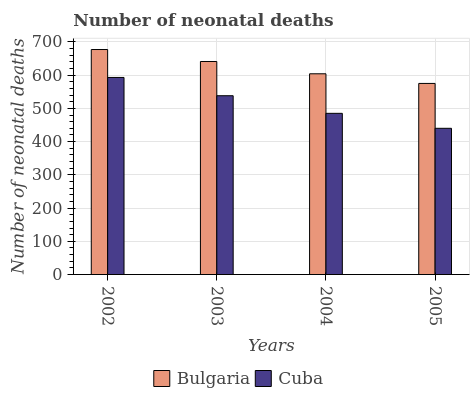}
\caption*{\textbf{Q2}: In how many years is the number of neonatal deaths in Cuba greater than 500? \\
\textbf{A}: 2
}
\end{subfigure}
\begin{subfigure}{.45\textwidth}
\centering
\includegraphics[scale=0.35]{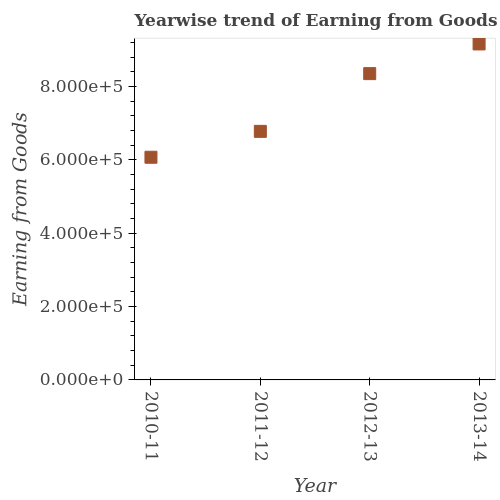}
\caption*{
\textbf{Q3}:  What is the difference between the highest and the second highest amount of earnings from goods?\\
\textbf{A}: $0.9\mathrm{e}+5$}
\end{subfigure}
\hspace{0.2cm}
\begin{subfigure}{.45\textwidth}
\centering
\includegraphics[scale=0.35]{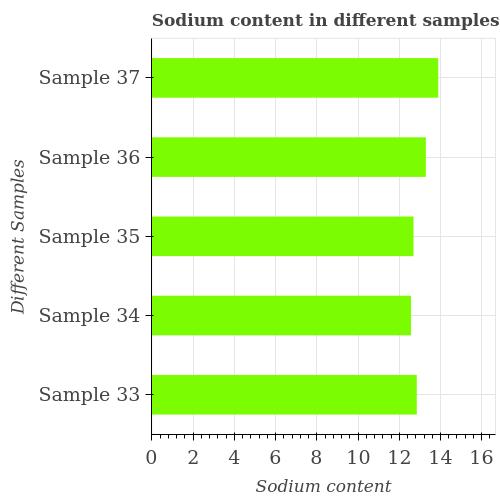}
\caption*{\textbf{Q4}: What is the ratio of the sodium content in Sample 37 to that in Sample 33?\\
\textbf{A}: 1.086
}
\end{subfigure}
\begin{subfigure}{.45\textwidth}
\includegraphics[scale=0.35]{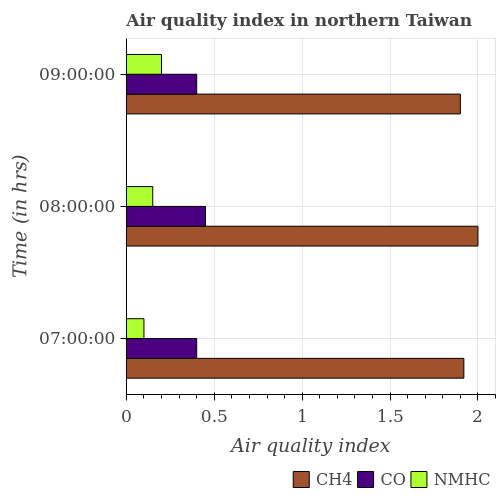}
\caption*{
\textbf{Q5}: What is the average air quality index value for NHMC per hour?\\
\textbf{A}: 0.167
}
\end{subfigure}
\hspace{0.2cm}
\begin{subfigure}{.45\textwidth}
\includegraphics[scale=0.35]{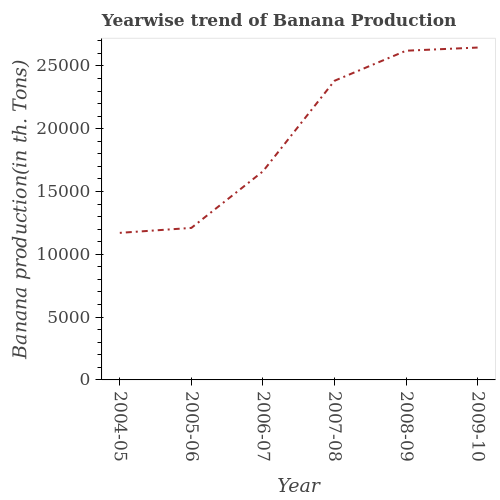}
\caption*{\textbf{Q6}: Does the amount of banana production monotonically increase over the years?\\
\textbf{A}: Yes
}
\end{subfigure}
\end{figure*}

\section{Question Templates}
\label{templates}
In this section, we present the 74 question templates which we have used for the question generation. Note that, not all question templates are applicable to each and every type of plot. Also depending on the context of the plot, the question varies from the template's surface form.
\begin{enumerate}
    \item \textbf{Structural Understanding} :
    \begin{enumerate}[1.]
        \item Does the graph contain any zero values?
        \item Does the graph contain grids ?
        \item Where does the legend appear in the graph ?
        \item How many legend labels are there?
        \item How are the legend labels stacked?
        \item How many $<$plural form of X\_label$>$ are there in the graph?
        \item How many $<$figure-type$>$s are there?
        \item How many different colored $<$figure-type$>$s are there?
        \item How many groups of $<$figure-type$>$s are there?
        \item Are the number of bars on each tick equal to the number of legend labels?
        \item Are the number of bars in each group equal?
        \item How many bars are there on the $i^{th}$ tick from the left?
        \item How many bars are there on the $i^{th}$ tick from the right?
        \item How many bars are there on the $i^{th}$ tick from the top?
        \item How many bars are there on the $i^{th}$ tick from the bottom?
        \item Are all the bars in the graph horizontal?
        \item How many lines intersect with each other?
        \item Is the number of lines equal to the number of legend labels?
    \end{enumerate}
    
    \item \textbf{Data Retrieval} :
    \begin{enumerate}[1.]
        \item What does the $i^{th}$ bar from the left in each group represent?
        \item What does the $i^{th}$ bar from the right in each group represent?
        \item What does the $i^{th}$ bar from the top in each group represent?
        \item What does the $i^{th}$ bar from the bottom in each group represent?
        \item What is the label of the $j^{th}$ group of bars from the left?
        \item What is the label of the $j^{th}$ group of bars from the top?
        \item Does the $<$Y\_label$>$ of/in $<$legend-label$>$ monotonically increase over the $<$plural form of X\_label$>$ ?
        \item What is the difference between two consecutive major ticks on the Y-axis ?
        \item Are the values on the major ticks of Y-axis written in scientific E-notation ?
        \item What is the title of the graph ?
        \item Does $<$legend\_label$>$ appear as one of the legend labels in the graph ? 
        \item What is the label or title of the X-axis ?
        \item What is the label or title of the Y-axis ?
        \item In how many cases, is the number of $<$figure\_type$>$ for a given $<$X\_label$>$ not equal to the number of legend labels ?
        \item What is the $<$Y\_value$>$ in/of $<i^{th}$ X\_tick$>$ ?
        \item What is the $<$Y\_value$>$ of the $i^{th}$ $<$legend\_label$>$ in $<i^{th}$ X\_tick$>$ ?
        \item Does the $<$Y\_label$>$ monotonically increase over the $<$plural form of X\_label$>$ ?
        \item Is the $<$Y\_label$>$ of/in $<$legend\_label1$>$ strictly greater than the $<$Y\_label$>$ of/in $<$legend\_label2$>$ over the $<$plural form of X\_label$>$ ?
        \item Is the $<$Y\_label$>$ of/in $<$legend\_label1$>$ strictly less than the $<$Y\_label$>$ of/in $<$legend\_label2$>$ over the $<$plural form of X\_label$>$ ?
    \end{enumerate}
    
    \item \textbf{Reasoning} :
    \begin{enumerate}[1.]
        \item Across all $<$plural form of X\_label$>$, what is the maximum $<$Y\_label$>$ ?
        \item Across all $<$plural form of X\_label$>$, what is the minimum $<$Y\_label$>$ ?
        \item In which $<$X\_label$>$ was the $<$Y\_label$>$ maximum ?
        \item In which $<$X\_label$>$ was the $<$Y\_label$>$ minimum ?
        \item Across all $<$plural form of X\_label$>$, what is the maximum $<$Y\_label$>$ of/in $<$legend\_label$>$ ?
        \item Across all $<$plural form of X\_label$>$, what is the minimum $<$Y\_label$>$ of/in $<$legend\_label$>$ ?
        \item In which $<$singular form of X\_label$>$ was the $<$Y\_label$>$ of/in $<$legend\_label$>$ maximum ?
        \item In which $<$singular form of X\_label$>$ was the $<$Y\_label$>$ of/in $<$legend\_label$>$ minimum ?
        \item What is the sum of $<$title$>$ ?
        \item What is the difference between the $<$Y\_label$>$ in $<i^{th}$x\_tick$>$ and $<j^{th}$x\_tick$>$ ?
        \item What is the average $<$Y\_label$>$ per $<$singular form of X\_label$>$ ?
        \item What is the median $<$Y\_label$>$ ?
        \item What is the total $<$Y\_label$>$ of/in $<$legend\_label$>$ in the graph?
        \item What is the difference between the $<$Y\_label$>$ of/in $<$legend\_label$>$ in $<i^{th}$x\_tick$>$ and that in $<j^{th}$x\_tick$>$ ?
        \item What is the difference between the $<$Y\_label$>$ of/in $<$legend\_label1$>$ in $<i^{th}$x\_tick$>$ and the $<$Y\_label$>$ of/in $<$legend\_label2$>$ in $<j^{th}$x\_tick$>$ ?
        \item What is the average $<$Y\_label$>$ of/in $<$legend\_label$>$ per $<$singular form of X\_label$>$ ?
        \item In the year $<i^{th}$x\_tick$>$, what is the difference between the $<$Y\_label$>$ of/in $<$legend\_label1$>$ and $<$Y\_label$>$ of/in $<$legend\_label2$>$ ?
        \item What is the difference between the $<$Y\_label$>$ of/in $<$legend\_label1$>$ and $<$Y\_label$>$ of/in $<$legend\_label2$>$ in $<i^{th}$x\_tick$>$ ?
        \item In how many $<$plural form of X\_label$>$, is the $<$Y\_label$>$ greater than $<$N$>$ units ?
        \item Do a majority of the $<$plural form of X\_label$>$ between $<i^{th}$ x\_tick$>$ and $<j^{th}$ x\_tick> (inclusive/exclusive)  have $<$Y\_label$>$ greater than N $<$units$>$ ?
        \item What is the ratio of the $<$Y\_label$>$ in $<i^{th}$ x\_tick$>$ to that in $<j^{th}$ x\_tick$>$ ?
        \item Is the $<$Y\_label$>$ in $<i^{th}$ x\_tick$>$ less than that in $<j^{th}$ x\_tick$>$ ?
        \item In how many $<$plural form of X\_label$>$, is the $<$Y\_label$>$ of/in $<$legend\_label$>$ greater than $<$N$>$ $<$units$>$?
        \item What is the ratio of the $<$Y\_label$>$ of/in $<$legend\_label1$>$ in $<i^{th}$ x\_tick$>$ to that in $<j^{th}$x\_tick$>$?
        \item Is the $<$Y\_label$>$ of/in $<$legend\_label$>$ in  $<i^{th}$ x\_tick$>$ less than that in $<j^{th}$ x\_tick$>$ ?
        \item Is the difference between the $<$Y\_label$>$ in $<i^{th}$x\_tick$>$ and $<j^{th}$x\_tick$>$ greater than the difference between any two $<$plural form of X\_label$>$ ?
        \item What is the difference between the highest and the second highest $<$Y\_label$>$ ?
        \item Is the sum of the $<$Y\_label$>$ in $<i^{th}$x\_tick$>$ and $<(i+1)^{th}$x\_tick$>$ greater than the maximum $<$Y\_label$>$ across all $<$plural form of X\_label$>$ ?
        \item What is the difference between the highest and the lowest $<$Y\_label$>$ ?
        \item In how many $<$plural form of X\_label$>$, is the $<$Y\_label$>$ greater than the average $<$Y\_label$>$ taken over all $<$plural form of X\_label$>$ ?
        \item Is the difference between the $<$Y\_label$>$ of/in $<$legend\_label1$>$ in $<i^{th}$x\_tick$>$ and $<j^{th}$x\_tick$>$ greater than the difference between the $<$Y\_label$>$ of/in $<$legend\_label2$>$ in $<i^{th}$x\_tick$>$ and $<j^{th}$x\_tick$>$ ?
        \item What is the difference between the highest and the second highest $<$Y\_label$>$ of/in $<$legend\_label$>$ ?
        \item What is the difference between the highest and the lowest $<$Y\_label$>$ of/in $<$legend\_label$>$ ?
        \item In how many $<$plural form of X\_label$>$, is the $<$Y\_label$>$ of/in $<$legend\_label$>$ greater than the average $<$Y\_label$>$ of/in $<$legend\_label$>$ taken over all $<$plural form of X\_label$>$ ?
        \item Is it the case that in every $<$singular form of X\_label$>$, the sum of the $<$Y\_label$>$ of/in $<$legend\_label1$>$ and $<$legend\_label2$>$ is greater than the $<$Y\_label$>$ of/in $<$legend\_label3$>$ ?
        \item Is the sum of the $<$Y\_label$>$ of/in $<$legend\_label1$>$ in $<i^{th}$x\_tick$>$ and $<j^{th}$x\_tick$>$ greater than the maximum $<$Y\_label$>$ of/in $<$legend\_label2$>$ across all $<$plural form of X\_label$>$?
        \item Is it the case that in every $<$singular form of X\_label$>$, the sum of the $<$Y\_label$>$ of/in $<$legend\_label1$>$ and $<$legend\_label2$>$ is greater than the sum of $<$Y\_label$>$ of $<$legend\_label3$>$ and $<$Y\_label$>$ of $<$legend\_label4$>$ ?
    \end{enumerate}
    
\end{enumerate}

\end{document}